\documentclass[10pt]{article}

\usepackage{hltnaacl03}
\usepackage{times}
\usepackage{latexsym}
\usepackage{epsfig}
\usepackage{xspace}
\newcommand{\epsfscaledbox}[2]{\centerline{\psfig{figure=#1,width=#2}}}
\newcommand{\omt}[1]{}
\newcommand{\bibsnip}{\vspace*{-.11in}}
\newcommand{\proc}{Proc.\xspace}

\newcommand{\Lattice}{Lattice\xspace}
\newcommand{\lattice}{lattice\xspace}
\newcommand{\lattices}{lattices\xspace}
\newcommand{\slotlat}{slotted \lattice}
\newcommand{\slotlats}{slotted \lattices}
\newcommand{\template}[1]{{\sf #1}}

\newenvironment{frameit}[1]
  {\begin{tabular}{|p{#1}|}\hline}{\\\hline\end{tabular}}

\title{\vspace{-75pt}
{\normalsize {\it \hfill Proceedings of HLT/NAACL 2003}} \\ \mbox{}\\Learning to Paraphrase: An Unsupervised Approach Using Multiple-Sequence Alignment}

\author{Regina Barzilay \and Lillian Lee \\ 
Department of Computer Science \\
Cornell University \\
Ithaca, NY 14853-7501 \\
\{regina,llee\}@cs.cornell.edu}

\date{}

\begin{document}
\maketitle
\begin{abstract}
  We address the text-to-text generation problem of sentence-level paraphrasing
  --- a phenomenon distinct from and more difficult than word- or phrase-level
  paraphrasing.  Our approach applies {\em multiple-sequence alignment} to
  sentences gathered from unannotated comparable corpora: it learns a set of
  paraphrasing patterns represented by {\em word lattice} pairs and
  automatically determines how to apply these patterns to rewrite new
  sentences.  The results of our evaluation experiments show that the system
  derives accurate paraphrases, outperforming baseline systems.
\end{abstract}

\section{Introduction}

\begin{quote}
{\em This is a late parrot! It's a stiff! Bereft of life, it rests in
peace! If you hadn't nailed him to the perch he would be pushing up
the daisies! Its metabolical processes are of interest only to
historians! It's hopped the twig! It's shuffled off this mortal coil!
It's rung down the curtain and joined the choir invisible! This is
an EX-PARROT!} --- Monty Python, ``Pet Shop'' 
\end{quote}

A mechanism for automatically generating multiple paraphrases of a
given sentence would be of significant practical import for
text-to-text generation systems.  Applications include summarization
\cite{Knight&Marcu:2000a} and rewriting
\cite{Chandrasekar+Srinivas:97a}: both could employ such a mechanism
to produce candidate sentence paraphrases
that other
system components  would filter for length, sophistication level, and
so forth.\footnote{Another interesting application,
  somewhat tangential to generation, would be
to expand existing corpora by providing several versions of their
component sentences.  
This could, for example, aid machine-translation evaluation, where it has
become common to evaluate systems by comparing their output against a bank of
several reference translations for the same sentences
\cite{Papineni&al:2002a}.
See \newcite{Bangalore&Murdock&Riccardi:2002a} and
\newcite{Barzilay&Lee:2002a} for other uses of such data.}
Not surprisingly, therefore,  
paraphrasing has been a focus of generation
research for quite some time
\cite{McKeown:79a,Meteer+Shaked:88a,Dras:1999a}. 

One might initially suppose that sentence-level paraphrasing is simply the
result of word-for-word or phrase-by-phrase substitution applied in a domain-
and context-independent fashion.  However, in studies of paraphrases across
several domains
\cite{Iordanskaja&Kittredge&Polguere:1991a,Robin-phd,McKeown&Kukich&Shaw:1994a},
this was generally not the case.
For instance, consider the following two sentences (similar to
examples found in  \newcite{Smadja&McKeown:1991a}):
  \begin{center}
    \begin{frameit}{0.9\columnwidth}
    {\small    After the latest Fed rate cut, stocks rose across the board.}
      \\\hline
      {\small Winners strongly outpaced losers after Greenspan cut
      interest rates again.}
    \end{frameit}
  \end{center}
  Observe that ``Fed'' (Federal Reserve) and ``Greenspan'' are interchangeable
  only in the domain of US financial matters.  Also, note that one cannot draw
  one-to-one correspondences between single words or phrases.  For instance,
  nothing in the second sentence is really equivalent to ``across the board'';
  we can only say that the entire clauses ``stocks rose across the board'' and
  ``winners strongly outpaced losers'' are paraphrases.  This evidence suggests
  two consequences: (1) we cannot rely solely on generic domain-independent
  lexical resources for the task of paraphrasing, and (2) {\em sentence-level}
  paraphrasing is an important problem extending beyond that of paraphrasing
  smaller lexical units.
  
  {\em Our work presents a novel knowledge-lean algorithm that uses {\em
      multiple-sequence alignment} (MSA) to {\em learn} to generate
    sentence-level paraphrases essentially from unannotated corpus data alone.}
  In contrast to previous work using MSA for generation
  \cite{Barzilay&Lee:2002a}, we need neither parallel data nor explicit
  information about sentence semantics.  Rather, we use two {\em comparable
    corpora}, in our case, collections of articles produced by two different
  newswire agencies about the same events.  The use of related corpora is key:
  we can capture paraphrases that on the surface bear little resemblance but
  that, by the nature of the data, must be descriptions of the same
  information.  Note that we also acquire paraphrases from each of the
  individual corpora; but the lack of clues as to sentence equivalence in
  single corpora means that we must be more conservative, only selecting as
  paraphrases items that are structurally very similar.
  
  Our approach has three main steps.  First, working on each of the comparable
  corpora separately, we compute {\em \lattices} --- compact graph-based
  representations --- to find commonalities within (automatically derived)
  groups of structurally similar sentences.  Next, we identify pairs of
  lattices from the two different corpora that are paraphrases of each other;
  the identification process checks whether the lattices take similar
  arguments.  Finally, given an input sentence to be paraphrased, we match it
  to a lattice and use a paraphrase from the matched lattice's mate to generate
  an output sentence.  The key features of this approach are:

\noindent
\textbf{Focus on paraphrase generation.} In contrast to earlier work, we not
only extract paraphrasing rules, but also automatically determine which of the
potentially relevant rules to apply to an input sentence and produce a revised
form using them.

\noindent
\textbf{Flexible paraphrase types.} Previous approaches to paraphrase
acquisition focused on certain rigid types of paraphrases, for instance,
limiting the number of arguments.  In contrast, our method is not limited to a
set of {\it a priori}-specified paraphrase types.

\noindent
\textbf{Use of comparable corpora and minimal use of knowledge resources.}  In
addition to the advantages mentioned above, comparable corpora can be easily
obtained for many domains, whereas previous approaches to paraphrase
acquisition (and the related problem of phrase-based machine translation
\cite{Wang:1998a,Och&Tillman&Ney:1999a,Vogel&Ney:2000a}) required parallel
corpora.  We point out that one such approach, recently proposed by
\newcite{Pang+Knight+Marcu:03a}, also represents paraphrases by lattices,
similarly to our method, although their lattices are derived using parse
information.

Moreover, our algorithm does not employ knowledge resources such as parsers or
lexical databases, which may not be available or appropriate for all domains
--- a key issue since paraphrasing is typically domain-dependent.  Nonetheless,
our algorithm achieves good performance.

\section{Related work}
Previous work on automated paraphrasing has considered different levels of
paraphrase granularity.  Learning synonyms via distributional similarity has
been well-studied \cite{Pereira&Tishby&Lee:1993a,Grefenstette:94a,Lin:1998a}.
\newcite{Jacquemin:l999a} and \newcite{Barzilay&McKeown:01a} identify
phrase-level paraphrases, while \newcite{Lin&Pantel:2001a} and
\newcite{Shinyama&al:2002a} acquire structural paraphrases encoded as
templates.  These latter are the most closely related to the sentence-level
paraphrases we desire, and so we focus in this section on template-induction
approaches.

\newcite{Lin&Pantel:2001a} extract inference rules, which are related to
paraphrases (for example, \template{X wrote Y} implies \template{X is the
  author of Y}), to improve question answering.  They assume that {\em paths}
in dependency trees that take similar arguments (leaves) are close in meaning.
However, only two-argument templates are considered.
\newcite{Shinyama&al:2002a} also use dependency-tree information to extract
templates of a limited form (in their case, determined by the underlying
information extraction application).  Like us (and unlike Lin and Pantel, who
employ a single large corpus), they use articles written about the same event
in different newspapers as data.

Our approach shares two characteristics with the two methods just described:
pattern comparison by analysis of the patterns' respective arguments, and use
of non-parallel corpora as a data source.  However, {\em extraction} methods
are not easily extended to {\em generation} methods.  One problem is that their
templates often only match small fragments of a sentence.  While this is
appropriate for other applications, deciding whether to use a given template to
generate a paraphrase requires information about the surrounding context
provided by the entire sentence.

\newcommand{\slot}{slot\xspace}
\newcommand{\slots}{slots\xspace}
\newcommand{\findclusters}{Sentence clustering}
\newcommand{\families}{clusters\xspace}
\newcommand{\Families}{Clusters\xspace}
\newcommand{\family}{cluster\xspace}
\newcommand{\famlat}{\lattice}
\newcommand{\famlats}{\lattices}
\newcommand{\msg}{pattern\xspace}

\newcommand{\patterninformal}{pattern\xspace}
\newcommand{\patternsinformal}{patterns\xspace}
\newcommand{\Patterninformal}{Pattern\xspace}
\newcommand{\surprise}{surprise\xspace}
\newcommand{\backbone}{backbone\xspace}
\newcommand{\numtoken}{NUM}
\newcommand{\nametoken}{NAME}
\newcommand{\datetoken}{DATE}

\section{Algorithm}

\paragraph{Overview} We first sketch the algorithm's broad outlines. The subsequent subsections provide
more detailed descriptions of the individual steps.

The major goals of our algorithm are to learn: 
\begin{itemize}
\item  recurring {\patternsinformal} in the data, such as  \template{X
(injured/wounded) Y people, Z seriously}, where the capital letters 
represent variables; 
\item
pairings between such \patternsinformal that represent paraphrases, for
example, between the \patterninformal \template{X (injured/wounded) Y people,
Z of them seriously} and the \patterninformal \template{Y were
(wounded/hurt) by X, among them Z were in serious condition}.
\end{itemize}

\begin{figure}
\begin{center}
\epsfscaledbox{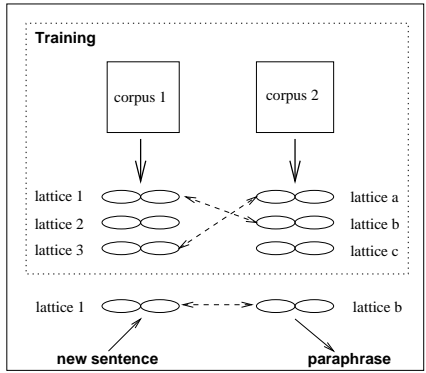}{2.2in}
\end{center}
\vspace*{-.2in}
\caption{\label{fig:arch} System architecture.}
\end{figure}

Figure~\ref{fig:arch} illustrates the main stages of our approach.  During
training, \patterninformal induction is first applied independently to the two
datasets making up a pair of {comparable corpora}.  Individual
\patternsinformal are learned by applying {\em multiple-sequence alignment} to
\families of sentences describing approximately similar events; these
\patternsinformal are represented compactly by {\em \lattices} (see Figure
\ref{fig:lattice}).  We then check for \lattices from the two different corpora
that tend to take the same arguments; these \lattice pairs are taken to be
paraphrase \patternsinformal.

Once training is done, we can generate paraphrases as follows: given the
sentence ``The \surprise bombing injured twenty people, five of them
seriously'', we match it to the lattice \template{X (injured/wounded) Y people,
  Z of them seriously} which can be rewritten as \template{Y were
  (wounded/hurt) by X, among them Z were in serious condition}, and so by
substituting arguments we can generate ``Twenty were wounded by the \surprise
bombing, among them five were in serious condition'' or ``Twenty were hurt by
the \surprise bombing, among them five were in serious condition''.

\begin{figure}
\newcounter{sentexample}\setcounter{sentexample}{1}
\newcommand{\sentex}[1]{{\footnotesize (\thesentexample)~#1 \stepcounter{sentexample}}}
\fbox{
\begin{minipage}{3in}
  \sentex{\textbf{A Palestinian suicide bomber blew himself up in} a southern
    city Wednesday, \textbf{killing} two other \textbf{people}
    \textbf{and wounding} 27.} \\
  \sentex{\textbf{A suicide bomber blew himself up in} the settlement of Efrat,
    on Sunday, \textbf{killing} himself \textbf{and injuring}
    seven people.} \\
  \sentex{\textbf{A suicide bomber blew himself up in} the coastal resort of
    Netanya on Monday, \textbf{killing} three other \textbf{people} 
    \textbf{and wounding} dozens more.} \\
  \sentex{\textbf{A Palestinian suicide bomber blew himself up in} a garden
    cafe on Saturday, \textbf{killing} 10 \textbf{people}  \textbf{and wounding}
    54.} \\
  \sentex{\textbf{A suicide bomber blew himself up in} the centre of Netanya on
    Sunday, \textbf{killing} three \textbf{people} as well as himself 
    \textbf{and injuring} 40. }
\end{minipage}
}
\caption{\label{fig:cluster} Five sentences (without date, number,
  and name substitution) from a \family of 49, similarities emphasized.  }
\end{figure}

\begin{figure*}
  \psfig{figure=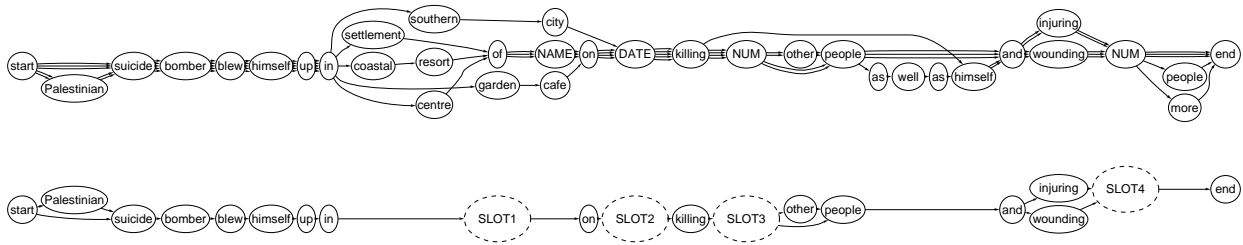,width=6.5in}
\caption{\label{fig:lattice} \Lattice and 
  \slotlat for the five sentences from Figure \ref{fig:cluster}.  Punctuation
  and articles removed for clarity.}
\end{figure*}

\subsection{\findclusters}

Our first step is to cluster sentences into groups from which to learn useful
patterns; for the multiple-sequence techniques we will use, this means that the
sentences within \families should describe similar events and have similar
structure, as in the sentences of Figure \ref{fig:cluster}.  This is
accomplished by applying hierarchical complete-link clustering to the sentences
using a similarity metric based on word n-gram overlap ($n=1,2,3,4$).  The only
subtlety is that we do not want mismatches on sentence details (e.g., the
location of a raid) causing sentences describing the same type of occurrence
(e.g., a raid) from being separated, as this might yield \families too
fragmented for effective learning to take place. (Moreover, variability in the
{\em arguments} of the sentences in a cluster is needed for our learning
algorithm to succeed; see below.)  We therefore first
replace all appearances of dates, numbers, and proper names\footnote{Our crude
  proper-name identification method was to flag every phrase (extracted by a
  noun-phrase chunker) appearing capitalized in a non-sentence-initial position
  sufficiently often.  }  with generic tokens.  \Families with fewer than ten
sentences are discarded.

\newcommand{\art}[1]{}
\newcommand{\monthtoken}{MONTH\xspace}
\newcommand{\mayseven}{\datetoken~\numtoken \xspace}
\newcommand{\palestinian}{\nametoken\xspace}
\newcommand{\southta}{\nametoken\xspace}
\newcommand{\marchnine}{\datetoken~\numtoken \xspace}
\newcommand{\jerusalem}{\nametoken\xspace}
\newcommand{\ipmasharon}{\nametoken\xspace}
\newcommand{\marchten}{\datetoken~\numtoken \xspace}
\newcommand{\saturday}{\datetoken\xspace}
\newcommand{\marchthirtyone}{\datetoken~\numtoken \xspace}
\newcommand{\afpsource}{\nametoken\xspace}
\newcommand{\jewish}{\nametoken\xspace}
\newcommand{\efratwwbbeth}{\nametoken\xspace}
\newcommand{\sunday}{\datetoken\xspace}
\newcommand{\juneeighteen}{\datetoken~\numtoken \xspace}
\newcommand{\fifteen}{\numtoken1\xspace}
\newcommand{\tuesday}{\datetoken\xspace}
\newcommand{\seven}{{\numtoken1}\xspace}
\newcommand{\eleven}{{\numtoken1}\xspace}
\newcommand{\fifty}{\numtoken2\xspace}
\newcommand{\eighteen}{\numtoken1\xspace}
\newcommand{\fortyeight}{\numtoken2\xspace}
\newcommand{\locone}{{in \art{a} crowded hall in \southta}\xspace}
\newcommand{\loconeshort}{in \art{a} crowded hall$\ldots$\xspace}
\newcommand{\loctwo}{into \art{a} crowded \jerusalem cafe [sic] \ipmasharon's residence\xspace}
\newcommand{\loctwoshort}{into \art{a} crowded $\ldots$ residence\xspace}
\newcommand{\locthree}{{in \art{the} \jewish
 settlement of \efratwwbbeth}\xspace}
\newcommand{\locthreeshort}{in \art{the} \jewish
 settlement $\ldots$ \xspace}
\newcommand{\locfour}{{aboard \art{a} crowded bus in \jerusalem}\xspace}
\newcommand{\locfourshort}{{aboard $\ldots$ \jerusalem}\xspace}
\newcommand{\synone}{injuring\xspace}
\newcommand{\syntwo}{wounding\xspace}

\subsection{Inducing \patternsinformal}

\newcommand{\simfn}{\textrm{sim}} \newcommand{\alphabet}{\Sigma}
\newcommand{\underscore}{\underline{~}} In order to learn \patternsinformal, we
first compute a {\em multiple-sequence alignment} (MSA) of the sentences in a
given \family.  Pairwise MSA takes two sentences and a scoring function giving
the similarity between words; it determines the highest-scoring way to perform
insertions, deletions, and changes to transform one of the sentences into the
other.  Pairwise MSA can be extended efficiently to multiple sequences via the
iterative pairwise alignment, a polynomial-time method commonly used in
computational biology \cite{Durbin+Eddy+al:98a}.\footnote{Scoring function:
  aligning two identical words scores 1; inserting a word scores -0.01, and
  aligning two different words scores -0.5 (parameter values taken from
  \newcite{Barzilay&Lee:2002a}).}  \omt{ $$\simfn(x,y) = 1 & $x = y$, $x \in
  \alphabet$; \cr -0.01 & exactly one of $x,y$ is $\underscore$~; \cr -0.5 &
  otherwise (mismatch)$$
  1 if the two words $x$ and $y$ are the same, -0.01 }
The results can be represented in an intuitive form via a word {\em \lattice}
(see Figure \ref{fig:lattice}), which compactly represents (n-gram) structural
similarities between the \family's sentences.

To transform \lattices into generation-suitable \patternsinformal requires some
understanding of the possible varieties of \lattice structures.  The most
important part of the transformation is to determine which words are actually
instances of arguments, and so should be replaced by {\em slots} (representing
variables).  The key intuition is that because the sentences in the \family
represent the same {\em type} of event, such as a bombing, but generally refer
to different {\em instances} of said event (e.g. a bombing in Jerusalem versus
in Gaza), areas of large variability in the \lattice should correspond to
arguments.

To quantify this notion of variability, we first formalize its opposite:
commonality.  We define {\em \backbone} nodes as those shared by more than 50\%
of the \family's sentences.  The choice of 50\% is not arbitrary --- it can be
proved using the pigeonhole principle that our strict-majority criterion
imposes a unique linear ordering of the backbone nodes that respects the word
ordering within the sentences, thus guaranteeing at least a degree of
well-formedness and avoiding the problem of how to order backbone nodes
occurring on parallel ``branches'' of the lattice.

Once we have identified the \backbone nodes as points of strong commonality,
the next step is to identify the regions of variability (or, in \lattice terms,
many parallel disjoint paths) between them as (probably) corresponding to the
arguments of the propositions that the sentences represent.  For example, in
the top of Figure \ref{fig:lattice}, the words ``southern city, ``settlement of
NAME'',``coastal resort of NAME'', etc.  all correspond to the location of an
event and could be replaced by a single {\slot}.
Figure \ref{fig:lattice} shows an example of a \lattice and the derived
\slotlat; we give the details of the slot-induction process in the Appendix.

\subsection{Matching \famlats}

Now, if we were using a parallel corpus, we could employ
sentence-alignment information to determine which lattices correspond
to paraphrases.  Since we do not have this information, we essentially
approximate the parallel-corpus situation by correlating information
from descriptions of (what we hope are) the same event occurring in
the two different corpora.

Our method works as follows.  Once \lattices for each corpus in our
comparable-corpus pair are computed, we identify \lattice paraphrase pairs,
using the idea that paraphrases will tend to take the same values as arguments
\cite{Shinyama&al:2002a,Lin&Pantel:2001a}. More specifically, we take a pair of
\lattices from different corpora, look back at the sentence clusters from which
the two lattices were derived, and compare the slot values of those
cross-corpus sentence pairs that appear in articles written on the {\em same
  day} on the same topic; we pair the \lattices if the degree of matching is
over a threshold tuned on held-out data.  For example, suppose we have two
(linearized) lattices \template{{slot1} bombed slot2} and \template{slot3 was
  bombed by slot4} drawn from different corpora.  If in the first lattice's
sentence cluster we have the sentence ``the plane bombed the town'', and in the
second lattice's sentence cluster we have a sentence written on the same day
reading ``the town was bombed by the plane'', then the corresponding lattices
may well be paraphrases, where \template{slot1} is identified with
\template{slot4} and \template{slot2} with \template{slot3}.

To compare the set of argument values of two lattices, we simply count their
word overlap, giving double weight to proper names and numbers and discarding
auxiliaries (we purposely ignore order because paraphrases can consist of word
re-orderings).

\subsection{Generating paraphrase sentences}

Given a sentence to paraphrase, we first need to identify which, if any, of our
previously-computed sentence \families the new sentence belongs most strongly
to. We do this by finding the best alignment of the sentence to the existing
\famlats.\footnote{ To facilitate this process, we add ``insert'' nodes between
  \backbone nodes; these nodes can match any word sequence and thus account for
  new words in the input sentence.  Then, we perform multiple-sequence
  alignment where insertions score \mbox{-0.1} and all other node alignments
  receive a score of unity.}  If a matching \famlat is found, we choose one of
its comparable-corpus paraphrase \lattices to rewrite the sentence,
substituting in the argument values of the original sentence.  This yields as
many paraphrases as there are lattice paths.

\section{Evaluation}
\label{sec:eval}

All evaluations involved judgments by native speakers of
English who were not familiar with the paraphrasing systems
under consideration.

\begin{figure*}
\epsfscaledbox{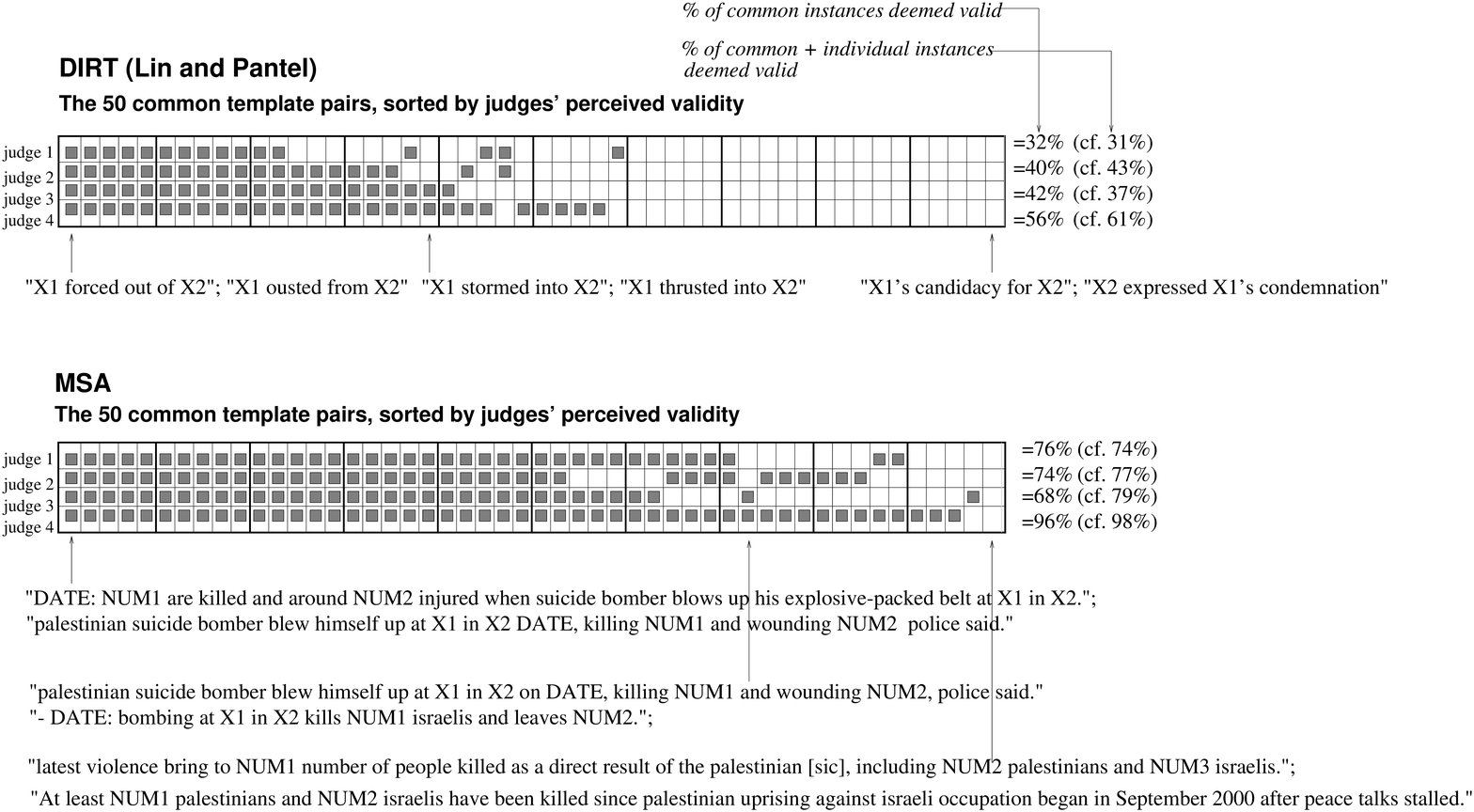}{6.4in}
\caption{\label{msa-dirt-accuracy} Correctness and agreement results.
Columns = instances; each grey box represents a judgment of ``valid''
for the instance.  For each method, a good, middling, and poor
instance is shown.  (Results separated by algorithm for clarity; the
blind evaluation presented instances from the two algorithms in random
order.)
} 
\end{figure*}

We implemented our system on a pair of comparable corpora consisting of
articles produced between September 2000 and August 2002 by the Agence
France-Presse (AFP) and Reuters news agencies.  Given our interest in
domain-dependent paraphrasing, we limited attention to 9MB of articles,
collected using a TDT-style document clustering system, concerning individual
acts of violence in Israel and army raids on the
Palestinian territories.  From this data (after removing 120 articles as a
held-out parameter-training set), we extracted 43 \slotlats from the AFP corpus
and 32 \slotlats from the Reuters corpus, and found 25 cross-corpus matching
pairs; since \lattices contain multiple paths, these yielded 6,534 template
pairs.\footnote{The extracted paraphrases are available at \texttt{http://www.cs.cornell.edu/Info/Projects/\\NLP/statpar.html}}

\subsection{Template Quality Evaluation}

Before evaluating the quality of the rewritings produced by our templates and
\lattices, we first tested the quality of a random sample of just the template
pairs.  In our instructions to the judges, we defined two {text units} (such as
sentences or snippets) to be paraphrases if one of them can generally be
substituted for the other without great loss of information (but not
necessarily vice versa).  \footnote{We switched to this ``one-sided''
  definition because in initial tests judges found it excruciating to decide on
  equivalence.
  Also, in applications such as summarization some information loss is
  acceptable.}  Given a pair of {\em templates} produced by a system, the
judges marked them as paraphrases if for many instantiations of the templates'
variables, the resulting text units were paraphrases.  (Several labelled
examples were provided to supply further guidance).

To put the evaluation results into context, we wanted to compare against
another system, but we are not aware of any previous work creating templates
precisely for the task of generating paraphrases.  Instead, we made a
good-faith effort to adapt the DIRT system \cite{Lin&Pantel:2001a} to the
problem, selecting the 6,534 highest-scoring templates it produced when run on
our datasets. (The system of \newcite{Shinyama&al:2002a} was unsuitable for
evaluation purposes because their paraphrase extraction component is too
tightly coupled to the underlying information extraction system.)  It is
important to note some important caveats in making this comparison, the most
prominent being that DIRT was not designed with sentence-paraphrase generation
in mind --- its templates are much shorter than ours, which may have affected
the evaluators' judgments --- and was originally implemented on much larger
data sets.\footnote{To cope with the corpus-size issue, DIRT was trained on an
  84MB corpus of Middle-East news articles, a strict superset of the 9MB we
  used.  Other issues include the fact that DIRT's output needed to be
  converted into English: it produces paths like ``N:of:N
  $\langle$tide$\rangle$ N:nn:N'', which we transformed into ``Y tide of X'' so
  that its output format would be the same as ours.  } The point of this
evaluation is simply to determine whether another corpus-based
paraphrase-focused approach could easily achieve the same performance level.

In brief, the DIRT system works as follows. Dependency trees are
constructed from parsing a large corpus.  Leaf-to-leaf paths are
extracted from these dependency trees, with the leaves serving as
slots.  Then, pairs of paths in which the slots tend to be filled by
similar values, where the similarity measure is based on the mutual
information between the value and the slot, are deemed to be
paraphrases.

We randomly extracted 500 pairs from the two algorithms' output sets.  Of
these, 100 paraphrases (50 per system) made up a ``common'' set evaluated by
all four judges, allowing us to compute agreement rates; in addition, each
judge also evaluated another ``individual'' set, seen only by him- or herself,
consisting of another 100 pairs (50 per system). The ``individual'' sets
allowed us to broaden our sample's coverage of the corpus.\footnote{Each judge
  took several hours at the task, making it infeasible to expand the sample
  size further.}
The pairs were presented in random order, and the judges were
not told which system produced a given pair.

As Figure~\ref{msa-dirt-accuracy} shows, our system outperforms the DIRT
system, with a consistent performance gap for all the judges of about 38\%,
although the absolute scores vary (for example, Judge 4 seems lenient).  The
judges' assessment of correctness was fairly constant between the full
100-instance set and just the 50-instance common set alone.

 In terms of agreement, the Kappa value (measuring pairwise agreement
discounting chance occurrences\footnote{One issue is that the Kappa
statistic doesn't account for varying difficulty among instances.  For
this reason, we actually asked judges to indicate for each instance
whether making the validity decision was difficult.  However, the
judges generally did not agree on difficulty.  Post hoc analysis
indicates that perception of difficulty depends on each judge's
individual ``threshold of similarity'', not just the instance itself.
}) on the common set was 0.54, which corresponds to moderate
agreement~\cite{Landis&Koch:1977a}.  Multiway agreement is depicted in
Figure~\ref{msa-dirt-accuracy} --- there, we see that in 86 of 100
cases, at least three of the judges gave the same correctness
assessment, and in 60 cases all four judges concurred.

\subsection{Evaluation of the generated paraphrases}

Finally, we evaluated the quality of the paraphrase sentences generated by our
system, thus (indirectly) testing all the system components: pattern selection,
paraphrase acquisition, and generation.  We are not aware of another system
generating sentence-level paraphrases.  Therefore, we used as a baseline a
simple paraphrasing system that just replaces words with one of their
randomly-chosen WordNet synonyms (using the most frequent sense of the word
that WordNet listed synonyms for). The number of substitutions was set
proportional to the number of words our method replaced in the same sentence.
The point of this comparison is to check whether simple synonym substitution
yields results comparable to those of our algorithm.  \footnote{ We chose not
  to employ a language model to re-rank either system's output because such an
  addition would make it hard to isolate the contribution of the paraphrasing
  component itself.  }

\begin{figure*}[htpb]\footnotesize
\hspace*{-.2in}
   \begin{tabular}{|l|l|}    \hline
      Original (1) & {\em The caller identified the bomber as Yussef Attala, 20, from the
      Balata refugee camp near Nablus.}                  \\\hline
      MSA  &  The caller named the bomber as 20-year old Yussef Attala from the
      Balata refugee camp near Nablus.                    \\\hline
      Baseline & The company placed the bomber as Yussef Attala, 20, from the
      Balata refugee camp near Nablus.    \\\hline \hline
      Original (2) & {\em A spokesman for the group claimed responsibility for the attack
      in a phone call to AFP in this northern West Bank town}. \\\hline
      MSA  & The attack in a phone call to AFP in this northern West Bank town
      was claimed by a spokesman of the group.                     \\\hline
      Baseline & \parbox[t]{6in}{A spokesman for the grouping laid claim
      responsibility for the onslaught in a phone call to AFP  
      in this northern West Bank town. } \\\hline
    \end{tabular}
    \caption{Example sentences and generated paraphrases. Both judges felt 
    MSA preserved the meaning of (1) but not (2), and that neither
    baseline paraphrase was meaning-preserving.}
    \label{fig:WordNet}
\end{figure*}

\newcommand{\results}[2]{#2\xspace} For this experiment, we randomly selected
20 AFP articles about violence in the Middle East published later than the
articles in our training corpus.  Out of 484 sentences in this set, our system
was able to paraphrase 59 (12.2\%).  (We chose parameters that optimized
precision rather than recall on our small held-out set.)  We found that after
proper name substitution, only seven sentences in the test set appeared in the
training set,\footnote{Since we are doing unsupervised paraphrase acquisition,
  train-test overlap is allowed.}  which implies that \lattices boost the
generalization power of our method significantly: from seven to 59 sentences.
Interestingly, the coverage of the system varied significantly with article
length.  For the eight articles of ten or fewer sentences, we paraphrased
60.8\% of the sentences per article on average, but for longer articles only
9.3\% of the sentences per article on average were paraphrased.  Our analysis
revealed that long articles tend to include large portions that are unique to
the article, such as personal stories of the event participants, which explains
why our algorithm had a lower paraphrasing rate for such articles.

All 118 instances (59 per system) were presented in random order to two judges,
who were asked to indicate whether the meaning had been preserved.  Of the
paraphrases generated by our system, the two evaluators deemed
\results{59-11}{81.4\%} and \results{59-13}{78\%}, respectively, to be valid,
whereas for the baseline system, the correctness results were
\results{59-18}{69.5\%} and \results{59-20}{66.1\%}, respectively. Agreement
according to the Kappa statistic was 0.6.  Note that judging full sentences is
inherently easier than judging templates, because template comparison requires
considering a variety of possible slot values, while sentences are
self-contained units.

Figure \ref{fig:WordNet} shows two example sentences, one where our MSA-based
paraphrase was deemed correct by both judges, and one where both judges deemed
the MSA-generated paraphrase incorrect.  Examination of the results indicates
that the two systems make essentially orthogonal types of errors. The baseline
system's relatively poor performance supports our claim that whole-sentence
paraphrasing is a hard task even when accurate word-level paraphrases are
given.

\section{Conclusions}

We presented an approach for generating sentence level
paraphrases, a task not addressed previously. Our method learns
structurally similar patterns of expression from data and identifies
paraphrasing pairs among them using a comparable corpus. A flexible 
pattern-matching procedure allows us to paraphrase an unseen sentence by
matching it to one of the induced patterns. Our approach
generates both lexical and structural paraphrases.

Another contribution is the induction of MSA lattices from non-parallel data.
Lattices have proven advantageous in a number of NLP contexts
\cite{Mangu&Brill&Stolcke:00a,Bangalore&Murdock&Riccardi:2002a,Barzilay&Lee:2002a,Pang+Knight+Marcu:03a},
but were usually produced from \mbox{(multi-)p}arallel data, which may not be
readily available for many applications.  We showed that word lattices can be
induced from a type of corpus that can be easily obtained for many domains,
broadening the applicability of this useful representation.

\vspace*{-.1in}

\section*{Acknowledgments}

{\footnotesize{
    We are grateful to many people for helping us in this work.  We thank
    Stuart Allen, Itai Balaban, Hubie Chen, Tom Heyerman, Evelyn Kleinberg,
    Carl Sable, and Alex Zubatov for acting as judges.  Eric Breck helped us
    with translating the output of the DIRT system.  We had numerous very
    useful conversations with all those mentioned above and with Eli Barzilay,
    Noemie Elhadad, Jon Kleinberg (who made the ``pigeonhole'' observation),
    Mirella Lapata, Smaranda Muresan and Bo Pang.  We are very grateful to
    Dekang Lin for providing us with DIRT's output.  We thank the Cornell NLP
    group, especially Eric Breck, Claire Cardie, Amanda Holland-Minkley, and Bo
    Pang, for helpful comments on previous drafts.  This paper is based upon
    work supported in part by the National Science Foundation under ITR/IM
    grant IIS-0081334 and a Sloan Research Fellowship.  Any opinions, findings,
    and conclusions or recommendations expressed above are those of the authors
    and do not necessarily reflect the views of the National Science Foundation
    or the Sloan Foundation.

\vspace*{-.2in}

\bibliographystyle{llee-fullname}

}
}

\section*{Appendix}

In this appendix, we describe how we  insert slots into  \lattices to
form \slotlats.

\begin{figure}[h]
\epsfscaledbox{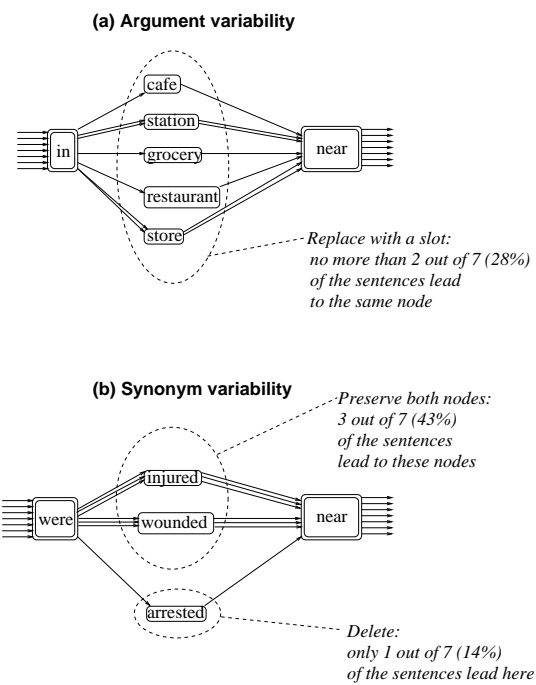}{2.8in}
\caption{\label{fig:variability} Simple seven-sentence examples of two types of
variability.  The double-boxed nodes are \backbone nodes; edges show
consecutive words in some sentence. The synonymy threshold (set to 30\%
in this example)
determines the type of variability. }
\end{figure}

Recall that the backbone nodes in our \lattices represent words appearing in
many of the sentences from which the lattice was built.  As mentioned above,
the intuition is that areas of high variability between backbone nodes may
correspond to arguments, or slots.  But the key thing to note is that there are
actually two different phenomena giving rise to multiple parallel paths: {\em
  argument variability}, described above, and {\em synonym variability}.  For
example, Figure \ref{fig:variability}(b) contains parallel paths corresponding
to the synonyms ``injured'' and ``wounded''.  Note that we want to {\em remove}
argument variability so that we can generate paraphrases of sentences with
arbitrary arguments; but we want to {\em preserve} synonym variability in order
to generate a variety of sentence rewritings.

To distinguish  these two situations, we analyze the {\em split
level} of \backbone nodes that begin regions with multiple paths. The
basic intuition is that there is probably more variability associated
with arguments than with
synonymy: for example, as datasets increase, the number of locations
mentioned rises faster than the number of synonyms appearing. We make
use of a 
{\em synonymy threshold} $s$ (set by held-out parameter-tuning
 to  30), as follows.
 
\begin{itemize}
\item If no more than $s$\% of all the edges out of a \backbone node
 lead to the same next node, we have high enough variability to
warrant inserting a {\slot} node. 
\item Otherwise, we incorporate reliable synonyms\footnote{While our original
    implementation, evaluated in Section~\ref{sec:eval}, identified only
    single-word synonyms, phrase-level synonyms can similarly be acquired by
    considering chains of nodes connecting backbone nodes.}  into the \backbone
  structure by preserving all nodes that are reached by at least $s$\% of the
  sentences passing through the two neighboring \backbone nodes.
\end{itemize} 
Furthermore, all \backbone nodes
labelled with our special generic tokens are
also replaced with \slot nodes, since they, too, probably represent arguments
(we condense adjacent \slots into one).  Nodes with in-degree lower than the
synonymy threshold are removed under the assumption that they probably
represent idiosyncrasies of individual sentences.  See Figure
\ref{fig:variability} for examples.

Figure \ref{fig:lattice} shows an example of a
\lattice and the  \slotlat derived via the process just described.

\end{document}